\documentclass{article}
\usepackage{spconf}
\usepackage[cmex10]{amsmath}
\usepackage{amsthm}
\usepackage{amssymb}
\usepackage{mathrsfs}
\usepackage{graphicx}
\usepackage{float}
\usepackage{array}
\usepackage{epstopdf}
\usepackage{multirow}

\newcommand{\argmax}{\arg\!\max}

\title{Face Recognition Using Scattering Convolutional Network}

\name{Shervin Minaee, Amirali Abdolrashidi and Yao Wang}
\address{ECE Department, NYU School of Engineering, USA
\\ \{shervin.minaee, abdolrashidi, yaowang\}@nyu.edu}

\allowdisplaybreaks[1]	
\begin{document}
%
\maketitle
\begin{abstract}
Face recognition has been an active research area in the past few decades. In general, face recognition can be very challenging due to variations in viewpoint, illumination, facial expression, etc. Therefore it is essential to extract features which are invariant to some or all of these variations. Here a new image representation, called scattering transform/network, has been used to extract features from faces. The scattering transform is a kind of convolutional network which provides a powerful multi-layer representation for signals.
After extraction of scattering features, PCA is applied to reduce the dimensionality of the data and then a multi-class support vector machine is used to perform recognition. 
The proposed algorithm has been tested on three face datasets and achieved a very high recognition rate.
\end{abstract}

\section{Introduction}
\label{sec:intro}
Face recognition is currently one of the most popular tasks in computer vision, which has many applications in authentication, security, imaging technology, etc. 
Because of the variations in the images of the same person such as changes in facial expression, pose, viewpoint and lighting conditions,  it could be quite challenging. A set of face images of one person with changes in facial expression and lighting are shown in Figure 1.
\begin{figure}[h]
\begin{center}
    \includegraphics [scale=0.20] {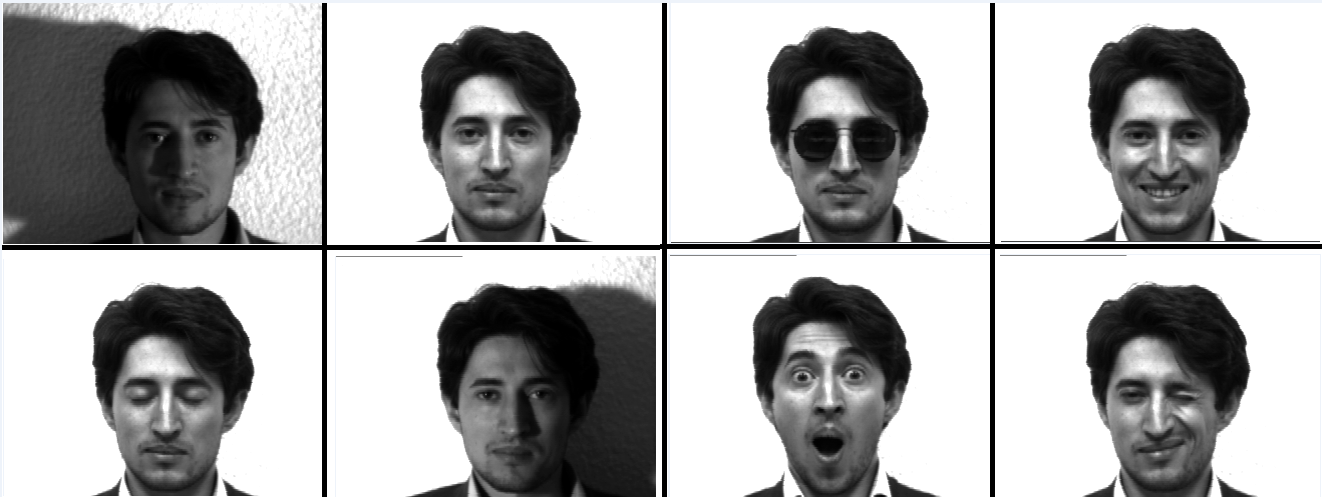}
\end{center}
  \vspace{-0.3cm}
  \caption{Eight face images with different facial expressions}
\end{figure}

Various algorithms have been proposed for face recognition during the past years. 
Despite the huge progress in the development of algorithms that are robust to variations in illumination, pose and  facial expression, there is still much room to explore for more reliable algorithms.
There have been various approaches for the face recognition problem during past years. 
Among those, Eigenface is one of popular approaches proposed in 1991 \cite{eigenface}. In this method, the face images are projected to a low-dimensional space using principal component analysis (PCA). These principal components which are derived from the training set are called Eigenfaces. 
In another work, known as Fisherfaces \cite{fisherface}, instead of using PCA, the author used a class-specific dimensionality reduction algorithm known as linear discriminant analysis (LDA). In this approach, two covariance matrices are defined, one measures the inter-class variation and the other evaluates the intra-class variation,
and the projection matrix is chosen as by maximizing the ratio of inter-class variation to intra-class variation.
Because LDA takes the class label into account, it provides more discriminative power than PCA.
Among more recent approaches, there have been a lot of works using sparse representation for face recognition. 
In \cite{wright}, Wright used sparse coding for face recognition which achieved a high accuracy rate.
In \cite{liu}, Liu proposed an extended version of sparse-coding where he added a non-negative constraint on the sparse coefficients. There are also more recent works such as the one in \cite{ASR}, \cite{RSR}, which uses adaptive or robust sparse representation for face recognition.
There are also many algorithms based on Bayesian framework for human face and pose recognition \cite{bay1}-\cite{bay2}.
There have also been a lot of works using hand-crafted features for face recognition such as using SIFT, HOG, Gabor wavelet and curvelet \cite{sift}-\cite{curve_PCA}.
In this paper, we propose an algorithm which uses a convolutional network called scattering transform/network which provides a multi-layer representation of the signal, and is invariant to translation, small deformation and rotation \cite{mallat1}, \cite{mallat2}.
After derivation of scattering features, their dimensionality is reduced using PCA and fed into a multi-class SVM to perform classification. This algorithm has been tested on three face databases and achieved very high accuracy rate.
The  scattering convolutional network can be used as a feature extractor for various object recognition tasks \cite{rahim1}-\cite{rahim3}, especially for problems with relatively small dataset \cite{med1}-\cite{med3}.

The rest of the paper is organized as follows. Section \ref{SectionII} provides a description of the scattering features, which are used in this work and also a brief overview of PCA. Section \ref{SectionIII} contains the explanation of the classification scheme. The experimental results and comparisons with other works are shown in Section \ref{SectionIV} and the paper is concluded in Section \ref{SectionV}.

\section{Features}
\label{SectionII}

Extracting good features is one of the most important steps in many of computer vision and object recognition tasks. As a result, a lot of research has been done in designing robust features for a variety of image classification tasks.
Good features should be invariant to the transformations which do not change object class.
Many image descriptors are proposed in the past two decades, including scale invariant feature transform (SIFT), histogram of oriented gradient (HOG), bag of words (BoW) \cite{sift_main}-\cite{BOW_main}.

Recently, unsupervised feature learning algorithms and deep convolutional neural networks have drawn a lot of attention and achieved state-of-the-art results in many computer vision problems \cite{Alex}. They are shown to provide more abstract and discriminative features which suit better for object recognition task.
In \cite{mallat1}, a wavelet-based multi-layer representation is proposed, which is similar to deep convolutional network, where instead of learning the filters and representation, it uses predefined wavelets \cite{mallat2}.
This network is especially useful for smaller datasets that is difficult to train a convolutional network from scratch, such as in medical and biomedical image analysis \cite{med1}-\cite{med4}.
This algorithm has been successfully applied to digit recognition, texture classification and audio classification problems and achieved state-of-the-art results \cite{mallat1}. 
It has also been used for some biometric recognition tasks such as iris, fingerprint and palmprint recognition \cite{iris}-\cite{finger}. 
In this paper, the application of scattering transform for face recognition is explored. 
The details of scattering features and their derivation are presented in the following section.

\subsection{Scattering Features}
Scattering transformation is a multi-layer representation recently proposed by Stephane Mallat. 
The scattering transformation can be designed such that it is invariant to a group of transformations such as translation, rotation, etc. Here a translation-invariant version is used.
It can be shown that the scattering coefficients of the first layer of the scattering network are similar to the SIFT descriptor, but the coefficients of the higher layers contain the high-frequency information lost in SIFT \cite{mallat1}. 
The scattering transform coefficients of each layer can be computed with a cascade of three operations: wavelet decompositions, complex modulus and  local averaging.

For a given signal $f(x)$ we can derive its scattering representation as follows. The first scattering coefficient is just the averaged signal which can be obtained by convolving the signal with the averaging filter $\phi_J$ as $f*\phi_J$. Then the scattering coefficients of the first layer are obtained by applying wavelet transforms of different scales and orientations, taking the magnitude of wavelet coefficients and convolving it by the averaging filter $\phi_J$ as shown below:
\begin{gather}
S_{1,J}(f(x)))= |f* \psi_{j_1,\lambda_1}|*\phi_J
\end{gather}
where $j_1$ and $\lambda_1$ denote the scale and orientation respectively.
By taking the magnitude of the wavelet, we can make these coefficients invariant to local translation.
On the other hand, some of the high-frequency information of the signal will be lost by averaging. 
We can recover some of the lost information by convolving the term $|f* \psi_{j_1,\lambda_1}|$ by another set of wavelets at scale $j_2<J$, taking the absolute value of wavelet followed by averaging as:
\begin{gather}
S_{2,J}(f(x)))= ||f* \psi_{j_1,\lambda_1}|*\psi_{j_2,\lambda_2}|*\phi_J
\end{gather}
It is enough to only calculate the coefficients for $j_1 >j_2 $, since $|f* \psi_{j_1,\lambda_1}|*\psi_{j_2,\lambda_2}$ is negligible for scales where $2^{j_1} \leq  2^{j_2}$.
We can continue this procedure to obtain the coefficients of the $k$-{th} layer of the scattering network as:
\begin{gather}
\underset{\ \ \ \ \ \ \ \ \ \ \ \ \ \ \ \ \ \ \ \ \ \ \ \ \ \ j_k<...<j_2<j_1<J, \ (\lambda_1,...,\lambda_k) \in \Gamma^k  }{S_{k,J}(f(x)))= ||f* \psi_{j_1,\lambda_1}|*...*\psi_{j_k,\lambda_k}|*\phi_J}
\end{gather}
It is easy to show that at the $k$-{th} layer of scattering representation will have $p^k {J \choose k}$ transformed signals where $p$ and $J$ denote the number of orientations and scales respectively.


Figures 2 and 3 denote the transformed images of the first and second layers of scattering transform for a sample face image. 
These images are derived by applying a filter bank of 5 different scales and 6 orientations.

\begin{figure}[h]
\begin{center}
    \includegraphics [scale=0.3] {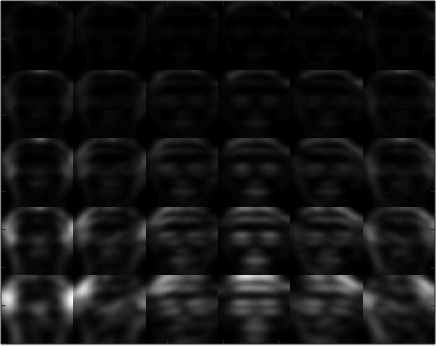}
\end{center}
  \vspace{-0.4cm}
  \caption{Images from the first layer of scattering transform}
\end{figure}
\begin{figure}[h]
\begin{center}
    \includegraphics [scale=0.35] {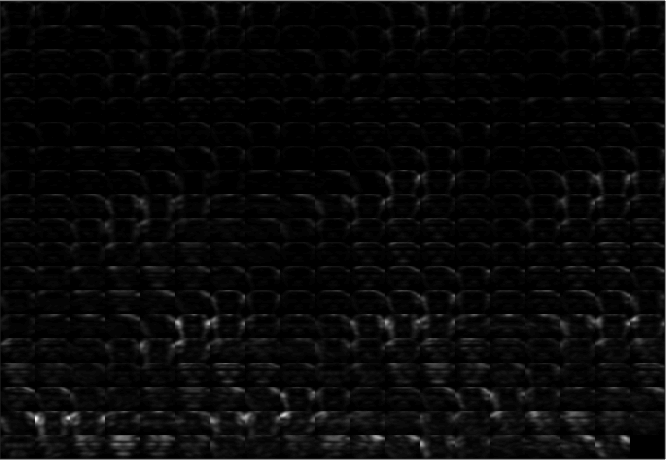}
\end{center}
\vspace{-0.2cm}
  \caption{Images from the second layer of scattering transform}
\end{figure}

As we can see, each image of the first layer is more sensitive to a specific orientation and scale.
To derive scattering features in our experiment, the scattering-transformed images of all layers up to $2$ are selected and the mean and variance of these images are calculated and used as scattering features which results in a vector $\textbf{f}_{\textbf{s}}$ of size $\sum_{k=0}^2{2p^k {J \choose k}}$.
The scattering features can be extracted either locally or globally.
Foreground segmentation schemes can also be used to detect important parts of the face, and extract features only around those regions  \cite{sp_sm}-\cite{sp_j}.

\subsection{Principal Component Analysis}
Dimensionality reduction algorithms are an essential part of most of today's object recognition algorithms to reduce the feature dimension such that the discriminant power of features are kept. Principal component analysis (PCA) is a powerful algorithm used for dimensionality reduction \cite{PCA}. Given a set of correlated variables, PCA transforms them to another domain such that the transformed variables are linearly uncorrelated. These linearly uncorrelated variables are called principal components. 

Let us assume $ \mathbf{x_i} \in \mathcal{R}^N $ denotes the features of the $i$-th image in the training set of $n$ samples, and we want to find the orthonormal matrix $\mathbf{W}= [ \mathbf{w_1}, \mathbf{w_2}, ..., \mathbf{w_N} ]$ to project the images as $\mathbf{y_i}= \mathbf{W}^T \mathbf{x_i}$ such that the variance of projected $\mathbf{y_i}$'s is maximized. One can show that the optimum $\mathbf{W}$ can be derived as: $\mathbf{W}^*=\argmax_{\mathbf{W}} || \mathbf{W}^T S_x \mathbf{W} ||$
such that $\mathbf{W}^T\mathbf{W}=\mathbf{I}$, where $S_x$ is the covariance of training images and is defined as $S_x= \sum_{i=1}^{n} (\mathbf{x_i}- \mu )(\mathbf{x_i}- \mu )^T$.
Then to reduce the dimension of the data to $m$, we can take the first $m$ eigenvectors associated with the $m$ largest eigenvalues of $S_x$ and project the data on them.

\section{Recognition Algorithm: Support Vector Machine}
\label{SectionIII}
After feature extraction, a classifier needs to be used to predict the label of each test image. Different machine learning algorithms can be used for classification.
In this work, support vector machine (SVM) \cite{SVM} is used to perform template matching. 
A brief overview of SVM in binary classification problem is provided here. 
Let us assume we are given a set of training data $(x_1,y_1)$, $(x_2,y_2)$, ..., $(x_n,y_n)$ and asked to classify them into two classes where $x_i \in \mathbf{R}^d $ is the feature vector and $y_i \in \{-1,+1\}$ is the class label. 
For linearly separable classes, two classes can be separated with a hyperplane $w.x+b=0$. Among all possible hyperplanes which can separate two classes, one reasonable choice is the one with the maximum margin. The maximum margin hyperplane can be derived by the following optimization problem:
\begin{equation}
\begin{aligned}
& \underset{w,b}{\text{minimize}}
& & \frac{1}{2} ||w||^2 \\
& \text{subject to}
& & y_i(w.x_i+b) \geq 1, \; i = 1, \ldots, n.
\end{aligned}
\end{equation} \\
It turns out solving this problem in dual domain is simpler than primal domain and since this problem is convex, the primal and dual solutions are the same. Then after derivation of Lagrange multipliers $\alpha_i$ in dual domain, we can derive the following hyperplane for classification: $f(x)= sign(\sum_{i=1}^{n} \alpha_i y_i x_i.x+b)$,
where $\alpha_i$ and $b$ are calculated by the SVM learning algorithm. Surprisingly, after solving this optimization problem, most of the $\alpha_i$'s are zero; therefore only the datapoints $x_i$ with nonzero $\alpha_i$ are important in the final classifier. These points are called support-vectors.
There is also a soft-margin version of SVM which allows for mislabeled examples by introducing a penalty term in the primal optimization problem with a penalty of $C$ times the degree of misclassification \cite{SVM}.

To derive the nonlinear classifier, one can map the data from input space into a higher-dimensional feature space $\mathcal{H}$ as: $x\rightarrow \phi(x)$, so that the classes are linearly separable in the feature space \cite{kernel_SVM}. If we assume there exists a kernel function where $k(x,y)= \phi(x).\phi(y)$, then we can use the kernel trick to construct nonlinear SVM by replacing the inner product $x.y$ with $k(x,y)$ which results in the following classifier:
\begin{equation}
f_n(x)= sign(\sum_{i=1}^{n} \alpha_i y_i K(x,x_i)+b)
\end{equation}

To derive multi-class SVM for a set of data with $M$ classes, we can train $M$ binary classifiers which can discriminate each class against all other classes, and to choose the class which classifies the test sample with the greatest margin. 
In another approach, we can train  a set of $M \choose 2$ binary classifiers, any of which separates one class from another one and to choose the class that is selected by the most classifiers. Other schemes have also been proposed for multi-class SVM.
For further detail and extensions to multi-class settings we refer the reader to \cite{multi_SVM}.

\section{Experimental results and analysis}
\label{SectionIV}
We have tested the proposed algorithm on three face databases, Yale Face Database,  Georgia Tech Face Database  and Extended Yale Face Database.
We first discuss about the parameter values of our algorithm and then present the results of this algorithm on different databases.

\subsection{Parameter Selection}
For each image, the scattering transform is applied up to two layers using a set of filters with 5 scales and 6 orientations, resulting in 391 transformed images. The mean and variance of each scatter-transformed images are used as features, which results in scattering features of dimension 782. The scattering features are derived using the software implementation provided by Mallat's group \cite{scat}. 
Then PCA is applied to all features and the first $K$ PCA features are used for recognition. Multi-class SVM is used for the classification. For SVM, we have used LIBSVM library \cite{libsvm}, and linear kernel is used in our implementation.

\subsection{Recognition Results on Three Databases}
\textbf{Yale Face Database:} This section presents the results of the proposed algorithm on Yale Face Database. This database contains 165 grayscale images of 15 individuals. There are 11 images per subject, one per different facial expression or configuration. We have performed recognition using multi-class SVM.
From each class, 6 images are used as training and the rest as test.
We repeat the experiment for 5 different sets of training images and report the average accuracy here.
Figure 4 demonstrates the recognition accuracy using different numbers of PCA features. For this case, the highest accuracy is achieved by using 200 PCA features which is around 93.1\%.
\begin{figure}[h]
\begin{center}
    \includegraphics [scale=0.43] {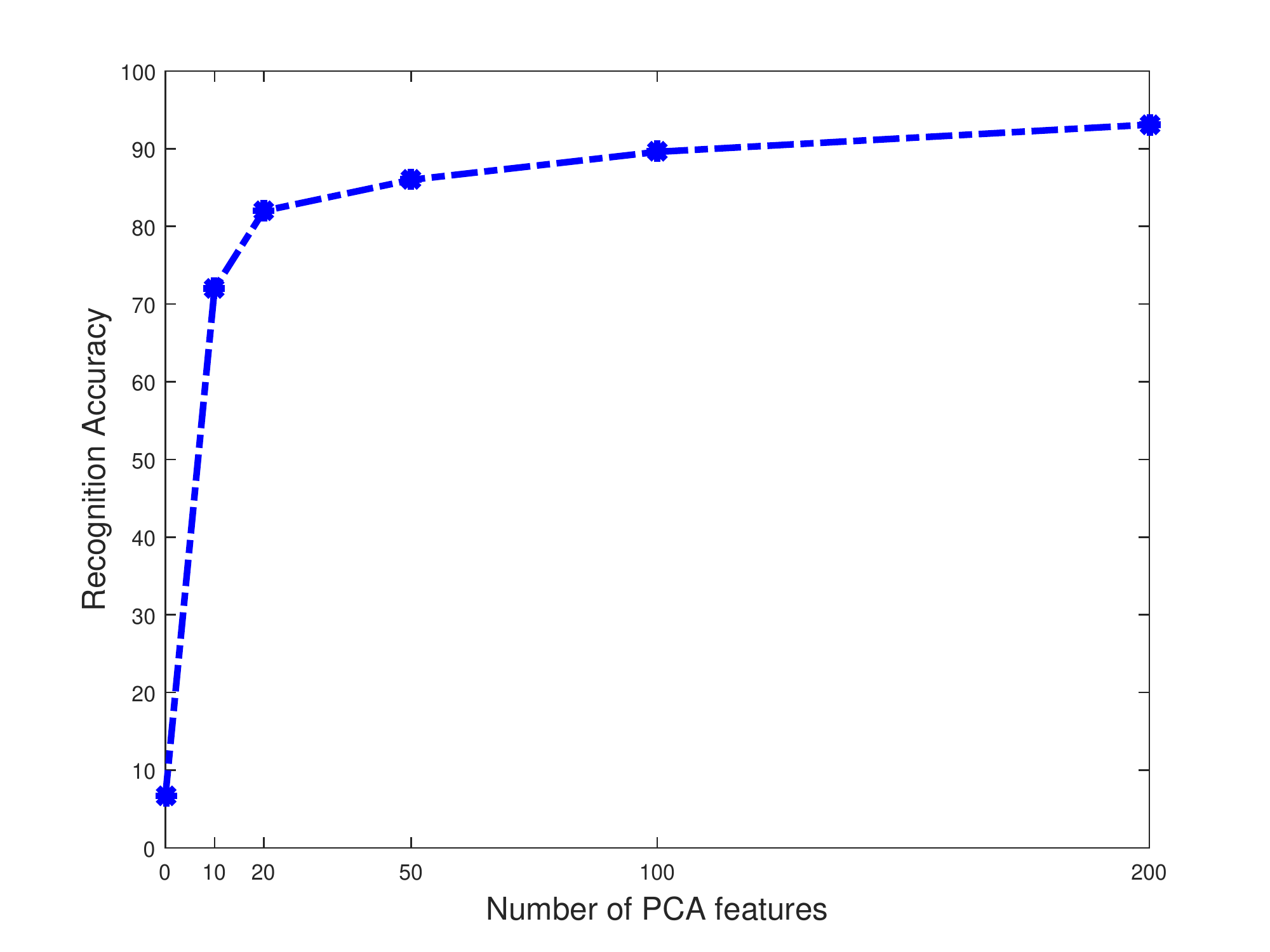}
\end{center}
  \vspace{-0.5cm}
  \caption{Recognition accuracy for Yale Face Database}
\end{figure}

\textbf{Georgia Tech Face Database:} Georgia Tech Face database contains images of 50 people taken in two or three sessions. For each subject in the database, there are 15 color images with cluttered background with different facial expressions, lighting conditions and scales.
For each person, 8 images are used as training and the rest as test.
Figure 5 shows the recognition accuracy using different numbers of PCA features and multi-class SVM. For this case, by using first 100 PCA features, an accuracy rate of around  90\% is achieved.
\begin{figure}[h]
\begin{center}
    \includegraphics [scale=0.43] {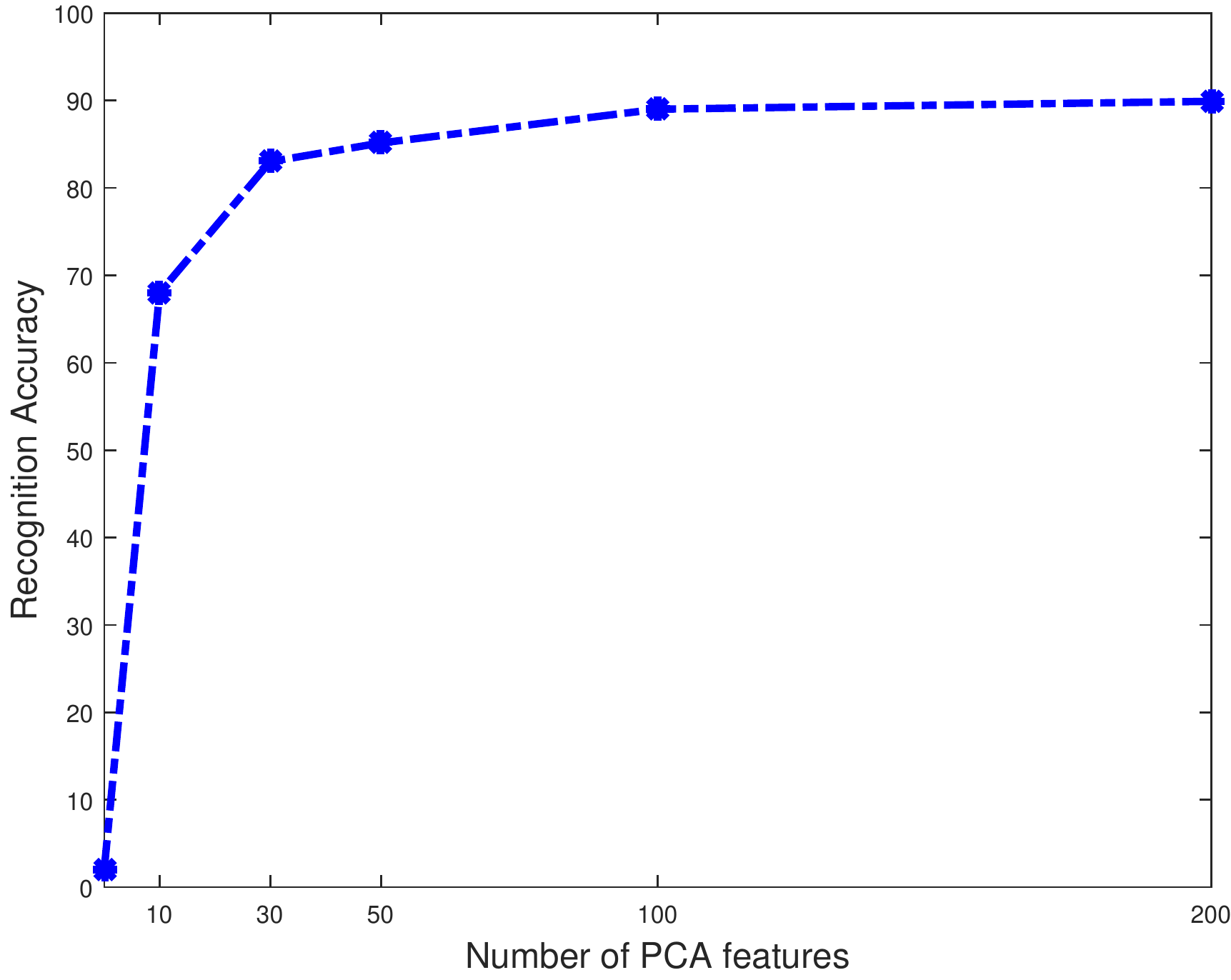}
\end{center}
  \vspace{-0.3cm}
  \caption{Recognition accuracy for GaTech Face Database}
\end{figure}

\textbf{Extended Yale Face Database:}
This database contains the frontal face images of 38 individuals. We used the cropped images, which were taken under varying illumination conditions and each subject has around 64 images.
We have used half of the images as the training and the rest as test.
Figure 6 shows the recognition accuracy using different numbers of PCA features. For this case, by using the first 200 PCA features, an accuracy rate of around 85\% will be achieved. 
\begin{figure}[h]
\begin{center}
    \includegraphics [scale=0.43] {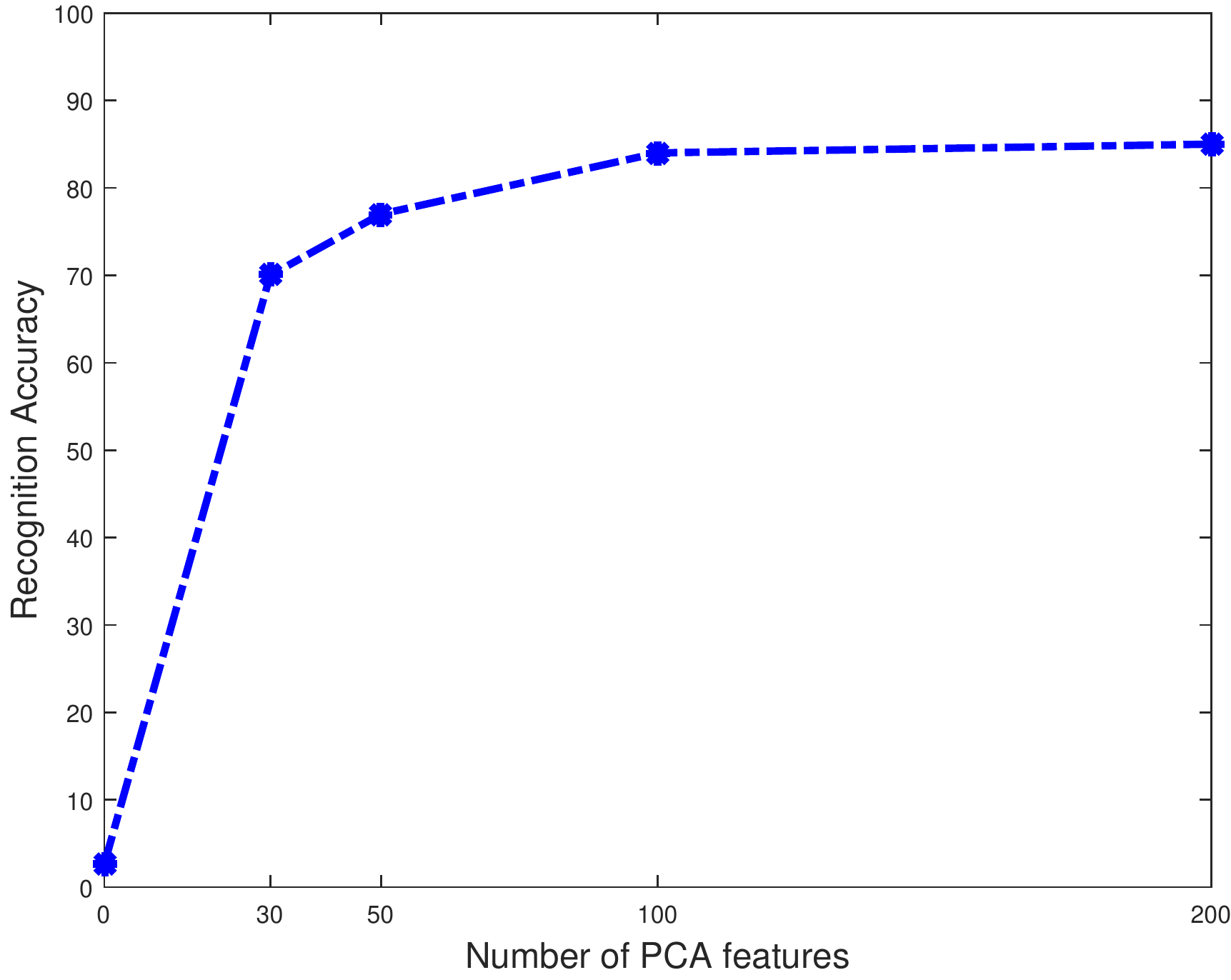}
\end{center}
  \vspace{-0.3cm}
  \caption{Recognition accuracy for Extended Yale Database}
\end{figure}

\subsubsection{Comparison With Previous Works}
Table 1 provides a comparison of the performance of the proposed scheme and that of 7 previous works on Yale face database. 
The recognition accuracy for EigenFaces 
is taken from the reported values in \cite{edgemap}.
We performed the experiment for 5 different sets of training images and reported the average accuracy below.
As it can be seen, the proposed approach outperforms the previous schemes on this database. \vspace{-0.2cm}
\begin{table} [h]
\centering
  \caption{Recognition accuracy comparison on Yale database}
  \centering
\begin{tabular}{|m{5.2cm}|m{2.2cm}|}
\hline
\ \ \ \ \ \ Face Recognition Method &  Recognition Rate\\
\hline
EdgeMap \cite{edgemap} & \ \ \ \ \ \ \  74\% \\
\hline
EigenFaces & \ \ \ \ \ \ \ 76\% \\
\hline
Sparse Representation Classifier \cite{ASR} & \ \ \ \ \ \ \ 82.34\% \\
\hline
EigenFaces w/o first 3 PCA & \ \ \ \ \ \ \ 84.7\% \\
\hline
Correlation based classifier \cite{ASR} & \ \ \ \ \ \ \ 85.1\% \\
\hline
Adaptive Sparse Representation \cite{ASR} & \ \ \ \ \ \ \ 88.06\% \\
\hline
Curvelet + PCA + LDA \cite{curve_PCA} & \ \ \ \ \ \ \ 92\% \\
\hline
Proposed algorithm & \ \ \ \ \ \ \ 93.1\% \\
\hline
\end{tabular}
\label{TblComp}
\end{table}



\vspace{-0.5cm}
\section{Conclusion}
\label{SectionV}
This paper proposed a face recognition algorithm using scattering convolutional network. Scattering features are locally invariant and carry a great deal of high-frequency information, which are lost in other descriptors such as SIFT and HOG. 
After feature extraction, PCA is applied to reduce dimensionality. Then multi-class SVM algorithm is used to perform recognition.
This algorithm has been tested on three well-known databases, and a high accuracy rate is achieved. 
The accuracy rate can be improved by using rotation-translation invariant scattering features, which is open for future research. 

\section*{Acknowledgments}
The authors would like to thank Mallat's research group for providing the software implementation of scattering transform.
We would also like to thank the CSIE group at NTU for providing LIBSVM software, and also research groups at Yale and Gatech for providing the face databases.

\end{document}